\documentclass[conference]{IEEEtran}
\IEEEoverridecommandlockouts
\usepackage{cite}
\usepackage{amsmath,amssymb,amsfonts}
\usepackage{algorithmic}
\usepackage{graphicx}
\usepackage{textcomp}
\usepackage{xcolor}

\usepackage[misc]{ifsym} 
\usepackage{caption}
\usepackage{booktabs}
\usepackage{url}
\usepackage{multirow}
\def\BibTeX{{\rm B\kern-.05em{\sc i\kern-.025em b}\kern-.08em
    T\kern-.1667em\lower.7ex\hbox{E}\kern-.125emX}}
\begin{document}

\title{Holistic and Historical Instance Comparison for Cervical Cell Detection\\}

\DeclareRobustCommand*{\IEEEauthorrefmark}[1]{%
  \raisebox{0pt}[0pt][0pt]{\textsuperscript{\footnotesize #1}}%
}

\author{
    \IEEEauthorblockN{
        Hao Jiang\IEEEauthorrefmark{1}, 
        Runsheng Liu\IEEEauthorrefmark{1},
        Yanning Zhou\IEEEauthorrefmark{2},
        Huangjing Lin\IEEEauthorrefmark{3}  and 
        Hao Chen\IEEEauthorrefmark{1, 4, 5, \Letter} \thanks{\textsuperscript{\Letter} Corresponding authors: Hao Chen. Email: jhc@cse.ust.hk} 
    }
    \IEEEauthorblockA{
        \IEEEauthorrefmark{1} Department of Computer Science and Engineering, \\The Hong Kong University of Science and Technology, Hong Kong, China\\
        \IEEEauthorrefmark{2} Tencent AI Lab, Shenzhen, China \\
        \IEEEauthorrefmark{3} Imsight AI Research Lab, Shenzhen, China \\
        \IEEEauthorrefmark{4} Department of Chemical and Biological Engineering, \\The Hong Kong University of Science and Technology, Hong Kong, China\\
        \IEEEauthorrefmark{5} HKUST Shenzhen-Hong Kong Collaborative Innovation Research Institute, Futian, Shenzhen, China
    }
}

\maketitle

\begin{abstract}
Cytology screening from Papanicolaou (Pap) smears is a common and effective tool for the preventive clinical management of cervical cancer, where abnormal cell detection from whole slide images serves as the foundation for reporting cervical cytology. However, cervical cell detection remains challenging due to 
1) hazily-defined cell types (e.g., ASC-US) with subtle morphological discrepancies caused by the dynamic cancerization process, i.e., cell class ambiguity,
and 2) imbalanced class distributions of clinical data may cause missed detection, especially for minor categories, i.e., cell class imbalance.  
To this end, we propose a holistic and historical instance comparison approach for cervical cell detection. 
Specifically,
we first develop a holistic instance comparison scheme enforcing both RoI-level and class-level cell discrimination. This coarse-to-fine cell comparison encourages the model to learn foreground-distinguishable and class-wise representations. To emphatically improve the distinguishability of minor classes, we then introduce a historical instance comparison scheme with a confident sample selection-based memory bank, which involves comparing current embeddings with historical embeddings for better cell instance discrimination. Extensive experiments and analysis on two large-scale cytology datasets including 42,592 and 114,513 cervical cells demonstrate the effectiveness of our method. The code is available at ~\url{https://github.com/hjiangaz/HERO}.
\end{abstract}


\section{Introduction}
Cervical cancer is one of the leading causes of cancer-related deaths, with approximately 604,127 confirmed cases and 341,831 deaths reported worldwide in 2020\cite{sung2021global}. Cytology screening, involving the identification of abnormal cells through the examination of thousands of cells under a microscope, is the primary approach for precancerous screening from Pap smears or liquid-based cytology specimens.
However, a typical cytology test usually requires experienced cytologists to spend 5-10 minutes on analyzing cytology characteristics under a microscope to identify abnormal cells \cite{lin2021dual}. Computational cytology has made significant progress in accelerating this screening process \cite{jiang2022deep}. Cell detection is usually regarded as the prerequisite step for identifying suspicious cells throughout the entire process \cite{jiang2020geometry,jiang2023donet,lin2021dual}.

\begin{figure*}[h!]
    \centering
\centerline{\includegraphics[width=0.9\textwidth]{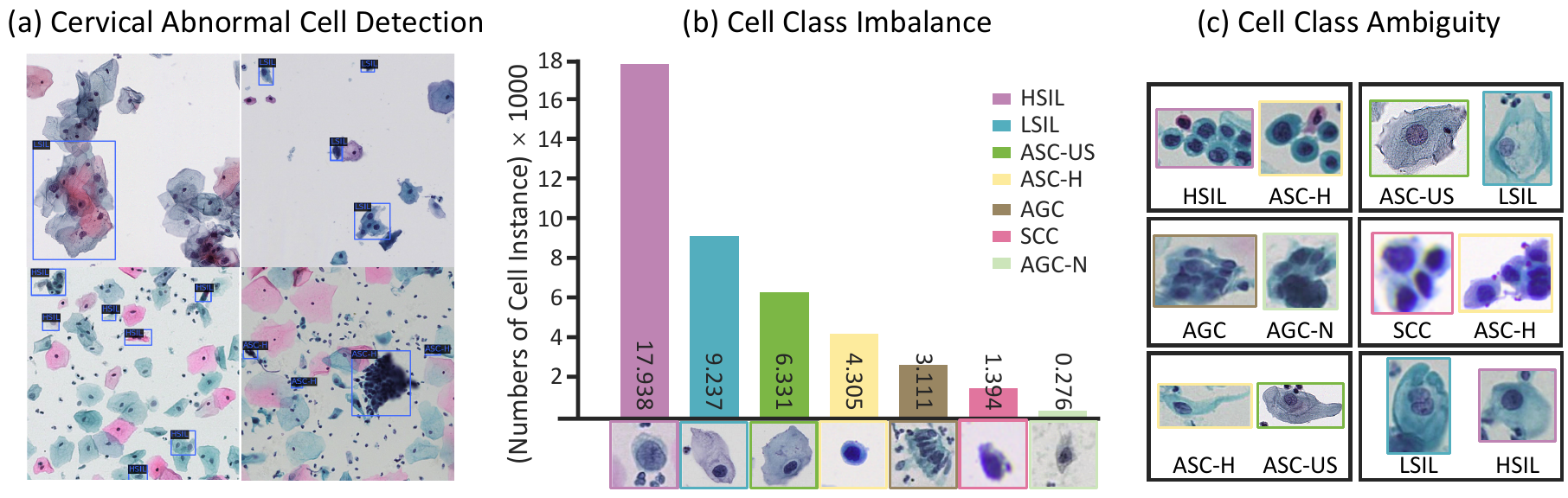}}
    \caption{Illustration of cervical cell detection. (a) Detected abnormal cells denoted by blue boxes; (b) \textbf{Cell class imbalance}: class imbalanced cell instance distribution; (c) \textbf{Cell class ambiguity}: cell class ambiguity with various appearances and morphologies. }
    \label{fig1}
    \vspace{-0.1in}
    \end{figure*}

Modern object detectors perform well in most natural scenes, but struggle to excel in this specific cervical cell detection task. Human Papillomavirus (HPV) invasion and infection processes are continuous and dynamic, with epithelial cells gradually progressing from low-grade lesions to cell carcinoma \cite{nayar2015bethesda}. Therefore, the morphological feature discrepancies between categories in adjacent stages  (e.g., HSIL and SCC) are subtle and indiscernible, resulting in class ambiguity (Fig. \ref{fig1}). Furthermore, the cervical cytology data is inherently under imbalanced class distributions due to screened positive candidates being mainly distributed in early stages of cancerization, which may lead to missing categories with fewer samples, i.e., SCC, AGC-N.

Previous works have emerged towards the cervical cell detection task. For example, Chai et al. \cite{chai2022deep} introduced a semi-supervised learning method to leverage unlabelled data for robust detection. Inspired by using surrounding cells as references in clinical practice, Liang et al. \cite{liang2022exploring} utilized two attention modules to explore contextual information. 
% Besides, Fei et al. \cite{fei2023robust} developed a two-level consistency learning to refine abnormal cell detection. 
Until recently, Liu et al. \cite{liu2022sample} observed the class imbalance issue in this task, while only focusing on class re-balancing. However, these studies only targeted specific issues, none of them tackled the intrinsic issue of cell indistinguishability caused by the blurred decision boundary between adjacent classes, which derives from the gradual progression of cancerization and is manifested in the morphological ambiguity.

To confirm the abnormal type in the case of cell class ambiguity, cytologists often retrieve TBS guidelines for deterministic reference and review previous samples for historical reference \cite{nayar2015bethesda}. Inspired by this, we propose a holistic instance comparison strategy, consisting of both deterministic and historical comparison within the same framework. Deterministic comparison aligns instances belonging to the same class in the RoI feature space and encourages current batch instances to be referenced against the ground truth (GT) instances, thereby learning the class distinguishability. For historical comparison, we introduce a confident sample selection-based memory bank in class-level comparison, which ensures that confident instances from minor classes are unbiasedly sampled and learnt in each batch.

The main contributions of this paper are as follows:
\begin{itemize}
\item To address the issues in cervical cell detection,  we present a holistic and historical instance comparison strategy to exploit comprehensive inter-cell instance discrimination.

\item  We introduce a RoI-level instance comparison module (RIC) and a class-level instance comparison module (CIC) with a confident sample selection-based memory bank to learn discriminative RoI and class representation.

\item Two large-scale cervical cell datasets containing 114,513 instance annotations, demonstrate our proposed method outperforms other state-of-the-art (SOTA) methods.

\end{itemize}

\begin{figure*}[h!]
    \centering
\centerline{\includegraphics[width=0.81\textwidth]{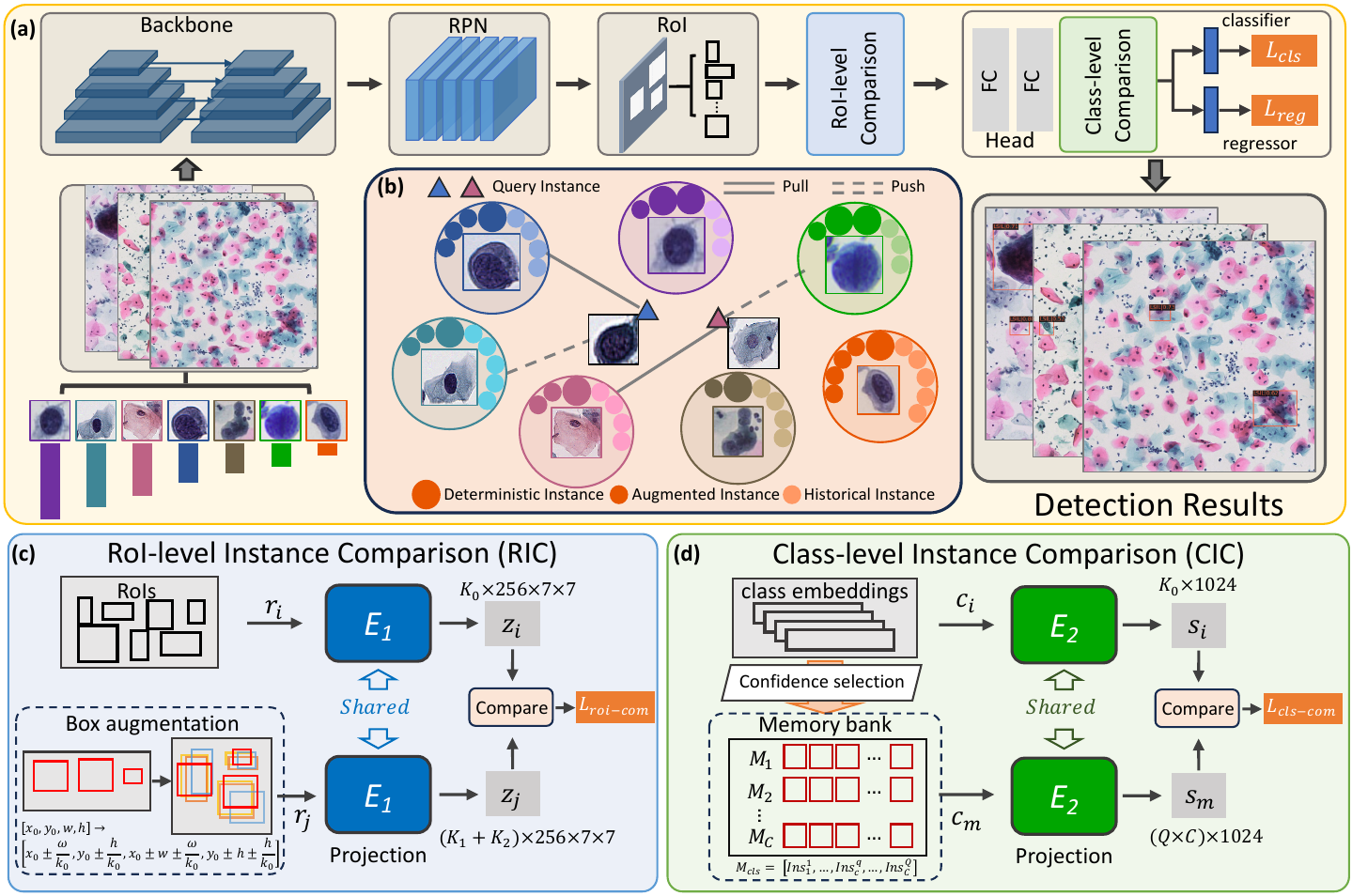}}
    \caption{Overview of the proposed Holistic and Historical Instance Comparison framework (a) for instance comparison (b). It consists of contrasting RoI features using RoI-level instance comparison module (RIC) with box augmentation (c), and contrasting class features by class-level comparison module  (CIC) with a confident sample selection-based memory bank (d).}
    \label{fig2}
    \vspace{-0.1in}
\end{figure*}

% \begin{table}[]
%     \centering
%     \small
%     % \footnotesize
%     \caption{Typical cytological abbreviation and terminology, more details refer to the Bethesda system \cite{nayar2015bethesda}.}
%     \begin{tabular}{p{0.065\textwidth}p{0.38\textwidth}}
%     % \hline
%     \toprule[1pt]
%         Abbr & Term\\ 
%         \cmidrule(lr){1-2}
%         ASC-US &Atypical squamous cells of undetermined significance \\ 
%         \cmidrule(lr){1-2}
%         LSIL & Low-grade squamous intraepithelial lesion \\ 
%         \cmidrule(lr){1-2}
%         ASC-H &  Atypical squamous cells cannot exclude HSIL \\ 
%         \cmidrule(lr){1-2}
%         HSIL & High-grade squamous intraepithelial lesion \\ 
%         \cmidrule(lr){1-2}
%         SCC & Squamous cell carcinoma \\ 
%         \cmidrule(lr){1-2}
%         AGC &Atypical glandular cells \\ 
%         \cmidrule(lr){1-2}
%         AGC-N & Atypical glandular cells, favor neoplastic \\ 
%         \bottomrule[1pt]
%     \end{tabular}
%     \label{tab0}
%      \vspace{-0.1in}
% \end{table}

\section{Methodology}

In this section, we first present the overview of proposed holistic and historical instance comparison approach  (Sec. \ref{s1-1}), then introduce the details of holistic instance comparison (Sec. \ref{s1-2}) and historical instance comparison (Sec. \ref{s1-3}), and finally the overall training schemes (Sec. \ref{sec1-4}).

\subsection{Framework Overview}\label{s1-1}

The proposed instance comparison method is based on the two-stage object detection framework \cite{ren2015faster}, as illustrated in Fig. \ref{fig2}. 
First, cytology images are fed into the backbone for feature extraction, followed by instance candidate generation through a Region Proposal Network (RPN). Then, these RoI candidates’ features are extracted by a projection head $E_{1}$. Then, we implement instance comparison on RoI feature maps to learn distinguishable instance features, through the RIC, which involves contrasting current RoI with GT embeddings. To explicitly address the class imbalance, we further introduce a confident sample selection-based memory bank in the CIC, which stores historical cell instances for each class with uniform sampling, thereby improving the generalizability in minority classes and avoid the domination of majority classes.

\subsection{Holistic Instance Comparison}\label{s1-2}

\noindent\textbf{RoI-level Instance Comparison (RIC).} Deterministic instance comparison on a large number of RoI and GT bounding boxes can increase the subtle inter-class discrepancy between cells and encourage high-quality RoI candidates generation.

RIC involves comparing RoI and GT embeddings through contrastive learning. It starts with assigning GTs and sampling proposals to obtain $bs \times k$ ($bs$ denotes batch size, $k=256$) RoI candidates with corresponding predicted classes $(0 \sim num.class)$, followed by filtering out the background class, obtaining $K_{0}$ RoIs. Then, the RoI feature extractor $E_{1}$ is used to generate RoI features $I$ with the size of $[K_{0}\times256\times7\times7]$. Then, for given $K_{1}$ GT boxes, we build a box augmentation method to enhance the diversity and amount of GT instances, enriching generalizable features in instance comparison. Given a box $b=\left[x_0, y_0, w, h\right]$, we randomly augment GT boxes as,
 \begin{equation}
B =\left[x_0 \pm \frac{w}{k_{0}}, y_0 \pm \frac{h}{k_{0}}, x_0+w \pm \frac{w}{k_{0}}, y_0+h \pm \frac{h}{k_{0}}\right],
 \end{equation}
obtaining ${K_{2}}$ augmented boxes. By performing similar operations through the shared feature extractor $E_{1}$, we get augmented GT features $J$ with the size of $[(K_{1}+K_{2})\times256\times7\times7]$. The next step involves supervised contrastive learning with positive ($I^{+}$) and negative ($I^{-}$) sample selection, where positive pairs from the same class and negative pairs from different classes, and GTs serve as current query batches $J$. 
 \begin{equation}
\mathcal{L}_{roi\_com}= - \sum_{j \in J} \frac{1}{|J|} \sum_{i^{+} \in I^{+}} \log \frac{\exp \left(\frac{Sim(z_j, z_{i^{+}}) }{ \tau_{roi}}\right)}{\sum_{i \in I} \exp \left(\frac{Sim(z_j \cdot z_i)}{ \tau_{roi}}\right)},
 \end{equation}
where $z_{i^{+}}$ represents the embedding of positive sample for the current query sample $z_{j}$. $Sim(z_{i},z_{j})=z_{i} \cdot z_{j} /(\left\|z_{i}\right\|\left\|z_{j}\right\|)$ is the function for calculating the similarity between two samples. $\tau_{roi}$ is a tunable temperature hyper-parameter. For the denominator, $I=I^{+}+I^{-}$ denotes the total number of positive and negative samples.

\noindent\textbf{Class-level Instance Comparison (CIC).}
To further improve class distinguishability, we design a class-level instance comparison module, which is located between the shared head and classification head to learn explicit instance class discrepancies. Specifically, we utilize historical class instance embeddings (detailed in Sec. \ref{s1-3}) and current batch class instance embeddings for instance comparison. First, RoI features are fed into the shared head (two fully connected layers) to obtain class features $I^{\prime}$ (with the size $[K_{0}\times1,024]$). Then, we conduct contrastive learning in a similar manner. Given the current class embedding $s_{i} \in I^{\prime}$ as query samples with a batches size of $|I^{\prime}|$, historical class embedding $s_m \in M = \{M^{+}, M^{-}\}$ including positive ($s_{m^{+}}$) and negative ($s_{m^{-}}$) samples. Thus, the instance class comparison loss function is formulated as,
 \begin{equation}
\mathcal{L}_{cls\_com}= - \sum_{i \in I^{\prime}} \frac{1}{|I^{\prime}|} \sum_{m^{+} \in M^{+}} \log \frac{\exp \left(\frac{Sim(s_i, s_{m^{+}}) }{ \tau_{cls}}\right)}{\sum_{m \in M} \exp \left(\frac{Sim(s_i \cdot s_m)}{ \tau_{cls}}\right)}.
 \end{equation}

\begin{figure*}[t!]
    \centering
    \centerline{\includegraphics[width=\textwidth]{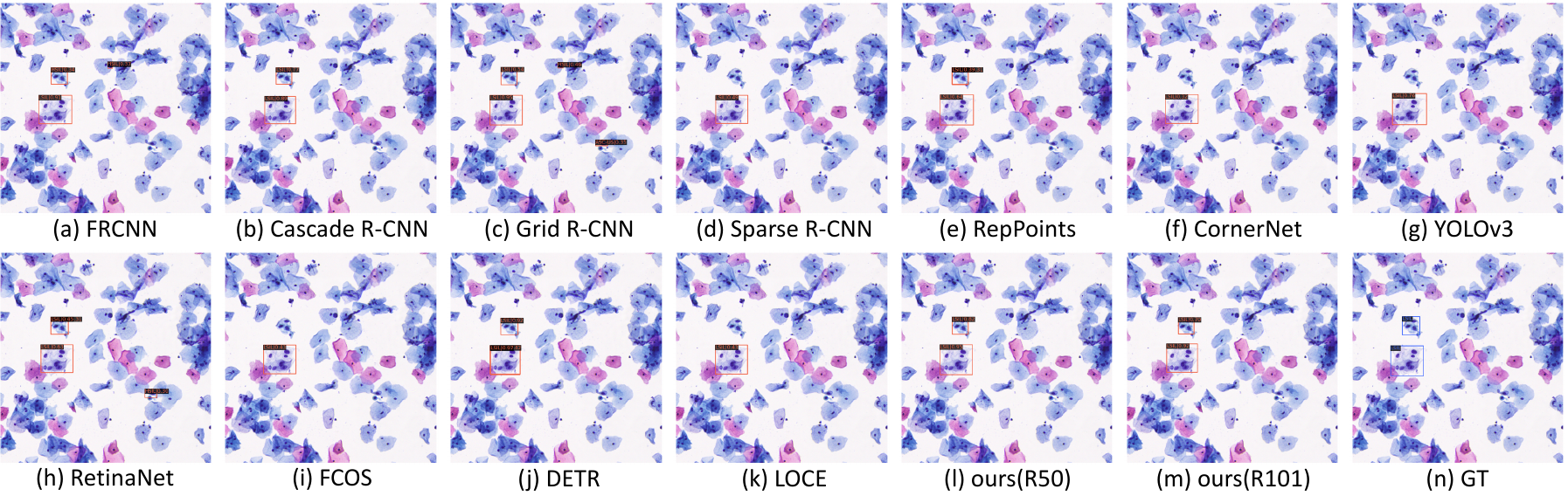}}
    \caption{Qualitative results of our methods and other SOTA methods. (a) FRCNN, (b) Cascade R-CNN, (c) Grid R-CNN, (d) Sparse R-CNN, (e) RepPoints, (f) CornerNet, (g) YOLOv3, (h) RetinaNet, (i) FCOS, (j) DETR, (k) LOCE, our method with R50 (l) and R101 (m), and (n) GT.}
    \label{fig4}
    % \vspace{-0.05in}
\end{figure*}

\begin{table*}[h!]
    \centering
    \small
    \caption{Quantitative detection results of our method and other SOTA methods with overall performance (AP50, AP75, AP, AR) and per-class performance. \textcolor{red}{Red} and \textcolor{blue}{blue} are first and second best results.}
    \label{tab4}
    \begin{tabular}{cccccccccccc}
    % \hline
    \toprule[1pt]
        Methods & AP50 $\uparrow$ & AP75 $\uparrow$& AP $\uparrow$& AR $\uparrow$ & ASC-US & LSIL & ASC-H & HSIL & SCC & AGC & AGC-N \\ 
        \cmidrule(lr){1-1}
        \cmidrule(lr){2-5}
        \cmidrule(lr){6-12}
        FRCNN (R50) \cite{ren2015faster} & 21.4 & 13.7 & 12.6 & 37.1 & 21.3 & 56.7 & 16.0 & 16.6 & 12.0 & 26.9 & 0.0 \\ 
        FRCNN (R101) \cite{ren2015faster} & 25.5& 18.0 & 16.6 & 41.0 & 32.2 & 60.8 & 21.0 & 18.7 & 12.8 & 33.1 & 0.0 \\
        Cascade R-CNN \cite{cai2018cascade} & 32.2 & 24.2 & 21.0 & 46.5 & 34.9 & 66.4 & \textcolor{blue}{34.5} & 23.2 & \textcolor{red}{32.7} & 33.8 & 0.0 \\
        Grid R-CNN \cite{lu2019grid} & 32.9 & 26.1 & 23.1 & \textcolor{red}{59.6} & 32.7 & 61.3 & 31.9 & 21.1 & 29.8 & 33.3 & \textcolor{red}{19.9} \\
        Sparse R-CNN \cite{sun2021sparse} & 20.6& 12.3 & 12.1 & 41.0 & 25.3 & 53.0 & 21.3& 14.6 & 2.1 & 26.2& 1.4 \\
        RepPoints \cite{yang2019reppoints} & 30.9 & 23.8 & 20.9 & 56.5 & 34.7 & 68.6 & 23.7 & 16.2 & 26.3 & 32.5 & 14.2 \\    
        CornerNet \cite{law2018cornernet} & 12.5 & 10.2 & 8.5 & 42.9 & 6.5 & 39.6 & 9.4 & 11.7 & 0.0 & 18.3 & 1.8 \\    
        YOLOv3 \cite{redmon2018yolov3}& 23.2 & 15.4 & 13.5 & 38.6 & 37.5 & 53.8 & 14.6 & 18.0 & 0.6 & 30.6 & 7.1 \\
        RetinaNet \cite{lin2017focal} & 28.1 & 21.4 & 18.3 & 52.9 & 38.1 & 61.3 & 18.6 & 20.6 & 5.3 & \textcolor{red}{36.9} & 15.7 \\
        FCOS \cite{tian2019fcos} & 25.5 & 16.6 & 15.5 & 50.8 & 26.2 & 48.8 & 19.4 & 15.8 & 28.3 & 29.0 & 10.8 \\
        DETR \cite{carion2020end} & 29.1 & 14.8 & 15.4 & 43.6 & 40.0	& 61.6 & 26.3 & 16.2 & 22.2 & 29.4 & 8.2 \\
        LOCE \cite{feng2021exploring} & 28.2 & 20.4 & 18.0 & \textcolor{blue}{59.1} & 27.6 & 55.2 & 22.5 & 19.4 & 24.5 & 32.1 & \textcolor{blue}{16.1} \\
        Ours (R50)& \textcolor{blue}{35.2} & \textcolor{blue}{28.3} & \textcolor{blue}{23.6} & 46.9 & \textcolor{blue}{47.4} & \textcolor{blue}{73.3} & \textcolor{red}{35.4} & \textcolor{red}{27.2} & 21.8 & 34.1 & 6.9 \\
        Ours (R101)& \textcolor{red}{36.8} & \textcolor{red}{28.4} & \textcolor{red}{23.8} & 47.5 & \textcolor{red}{52.6} & \textcolor{red}{78.3} & 34.1 & \textcolor{blue}{25.2} & \textcolor{blue}{30.0} & \textcolor{blue}{34.8} & 3.0 \\ 
        \bottomrule[1pt]
    \end{tabular}
    \vspace{-0.15in}
\end{table*}

\subsection{Historical Instance Comparison}\label{s1-3}
Although comparing current batch instances in CIC often leads to learning class discrimination to address the class ambiguity issue, it mainly encourages class discrepancies among majority class instances. This is because minority classes (e.g., SCC, AGC-N) have very low frequencies, compared to the overwhelming majority classes (e.g., LSIL, ASC-US). Therefore, we propose a confident sample selection-based memory bank, which not only increases the frequency of minority class instances to avoid class biased learning, but also increases the number of class instances for each batch training, improving the model's generalizability.

Specifically, we use a feature memory bank to store class instance features in the current batch, and reuse these features during the following training. 
The memory bank has a size of $[C\times Q]$, where $C$ is the number of instance categories ($C=7, 15$ in experiments) and $Q$ is the queue size per class.
% ($16, 80, 160$ in ablations). 
The memory bank $M_{cls}$ is defined as follows, 
\begin{equation}
    M_{cls}=\left[Ins_1^1, \ldots, Ins_c^q, \ldots, Ins_C^{Q}\right],
\end{equation}
where $Ins_c^{q}$ represents the $q_{th}$ class instance feature of category $c$ in memory bank $M_{cls}$. 

The memory bank is dynamically updated using the queue-based scheme. Confident sample selection is designed to ensure the quality of queued instances in $M_{cls}$, which enhances instance comparison, as
% \begin{equation}
    $l_{c} \geq \tau_{c}.$
% \end{equation}
It means comparing the predicted class score $l_{c}$ of the current sample with the corresponding class confidence score $\tau_{c}$, selecting confident samples, and adding them to the queue to update the memory bank $M_{cls}$. To enhance the stability of the memory bank and leverage the benefit of the confidence selection strategy, we perform the warm-up by training the baseline model, then implement our instance comparison after a few epochs.

\subsection{Overall Loss Function and Optimization}\label{sec1-4}
The framework can be trained with the overall objective $\mathcal{L}$,
 \begin{equation}
 \mathcal{L} = \lambda_{roi\_com} \mathcal{L}_{roi\_com} + \lambda_{cls\_com} \mathcal{L}_{cls\_com} + \mathcal{L}_{base},
  \label{eq:eq1}
  \end{equation}
where $\mathcal{L}_{roi\_com}$ denotes RoI comparison loss, $\mathcal{L}_{cls\_com}$ denotes class comparison loss. $\mathcal{L}_{base}$ supervises the baseline training, including RPN loss $\mathcal{L}_{RPN}$, regression loss $\mathcal{L}_{reg}$, and classification loss $\mathcal{L}_{cls}$ following previous settings \cite{ren2015faster}. $\lambda_{roi\_com}$ and $\lambda_{cls\_com}$ are trade-off controlling parameters,
% controlling the importance of each instance comparison, 
set as $\{1, 0.1\}$ in experiments.

\section{Experiments and Results}

\subsection{Dataset and Experiments Settings}

\begin{table*}[t!]
    \centering
    \small
    % \footnotesize
    \caption{Effectiveness of proposed Instance Comparison strategy. \textcolor{red}{Red} and \textcolor{blue}{blue} are first and second best results.}

    \begin{tabular}{ccccccccccccccc}
    % \hline
    \toprule[1pt]
        Base & RIC & Aug & CIC & AP50 $\uparrow$ & AP75 $\uparrow$& AP $\uparrow$& AR $\uparrow$& ASC-US & LSIL & ASC-H & HSIL & SCC & AGC & AGC-N \\ 
        \cmidrule(lr){1-4}
        \cmidrule(lr){5-8}
        \cmidrule(lr){9-15}
        \checkmark & ~ & ~ & ~ &21.4 & 13.7 & 12.6 & 37.1 & 21.3 & 56.7 & 16.0 & 16.6 & 12.0 & 26.9 & 0.0 \\ 
        \checkmark & \checkmark & ~ & ~ & 32.7 & 25.2 & 21.5 & 44.5 & 42.1 & \textcolor{blue}{74.5} & 32.5 & 24.6 & 19.3 & 33.1 & \textcolor{blue}{3.0} \\
        \checkmark & \checkmark & \checkmark & ~ & \textcolor{blue}{34.5} & \textcolor{blue}{27.4} & \textcolor{blue}{23.2} & \textcolor{red}{47.2} & \textcolor{blue}{46.3} & \textcolor{red}{74.7} & \textcolor{blue}{35.2} & \textcolor{blue}{26.4} & \textcolor{blue}{21.4} & \textcolor{red}{34.7} & \textcolor{blue}{3.0} \\
        \checkmark & \checkmark & \checkmark & \checkmark & \textcolor{red}{35.2} & \textcolor{red}{28.3} & \textcolor{red}{23.6} & \textcolor{blue}{46.9} & \textcolor{red}{47.4} & 73.3 & \textcolor{red}{35.4} & \textcolor{red}{27.2} & \textcolor{red}{21.8} & \textcolor{blue}{34.1} & \textcolor{red}{6.9} \\ 
        \bottomrule[1pt]
    \end{tabular}
    \label{tab1}
     \vspace{-0.15in}
\end{table*}

\noindent\textbf{Dataset.} 
To validate the effectiveness of the proposed method, we build two cervical cytology datasets, called CC-L and CC-S, containing 39,006 and 17,301 images ($1,200 \times 1,200$), respectively, cropped from cytology whole slides. Based on the Bethesda system \cite{nayar2015bethesda}, the CC-L dataset contains 114,513 box annotations. As shown in Tab. \ref{tab-data}, these instance annotations exhibit imbalanced distributions across 15 categories. To our knowledge, this is the largest cervical cytology dataset. In addition, we filtered out microbial categories to focus on specific intraepithelial lesions, leaving 42,592 instances and splitting the dataset into train, test, and validation for experiments.

\begin{table}[h!]
    \centering
    \small
    % \footnotesize
    \caption{The details of instance annotations and class distribution in CC-S and CC-L datasets.}
    \begin{tabular}{p{0.04\textwidth}p{0.045\textwidth}p{0.045\textwidth}p{0.045\textwidth}p{0.045\textwidth}p{0.055\textwidth}p{0.045\textwidth}}
    % \hline
    \toprule[1pt]
        Dataset & Category& Instance & Category& Instance& Category& Instance\\ 
        \cmidrule(lr){1-7}
        \multirow{3}{*}{CC-S}&ASCUS&6,631&LSIL&9,273&ASC-H&4,305\\
        &HSIL&17,938&SCC&1,394&AGC-N&276\\
        &AGC&3,111&&&&\\
        \cmidrule(lr){1-7}
        \multirow{5}{*}{CC-L}&ASCUS&22,056&LSIL&26,428&ASC-H&8,802\\
        &HSIL&28,110&SCC&1,756&AGC-N&256\\
         &AGC&6,533&ACTI&395&EMC&9,950\\
          &CC&2,510&Atrophy&453&FUNGI&1,023\\
          &Metapla&3,574&TV&2,274&AIS&402\\ 
        \bottomrule[1pt]
    \end{tabular}
    \label{tab-data}
     \vspace{-0.15in}
\end{table}

\noindent\textbf{Evaluation Metrics.}
For quantitative evaluation, we adopt commonly-used average precision (AP) and average recall (AR) across all classes over bounding box IoU thresholds ranging from 0.5 to 0.95, and the individual AP under the IoU threshold of 0.5 and 0.75, denoted by AP50 and AP75. We also report AP for each class under the IoU threshold of 0.5 to show the detection performance for each class.

\noindent\textbf{Implementation Details.}
We utilize the Faster R-CNN (FRCNN) \cite{ren2015faster} as baseline model. We use ResNet-50-based and ResNet-101-based FPN in all experiments. For training, we utilize SGD with 0.9 momentum as the optimizer and set the initial learning rate to 0.005. To ensure convergence, we train the network for 24 epochs, reducing the learning rate by a factor of 0.1 after 8 and 14 epochs. All experiments were conducted using 24GB NVIDIA GeForce RTX 3090 GPUs.

\subsection{Comparison with SOTA methods}

\noindent\textbf{Qualitative Results.}
We compare detection results from the CC-S dataset with 11 state-of-the-art detectors, including R-CNN-based models (FRCNN \cite{ren2015faster}, Cascade R-CNN \cite{cai2018cascade}, Grid R-CNN \cite{lu2019grid}, Sparse R-CNN \cite{sun2021sparse}), point-based models (RepPoints \cite{yang2019reppoints}, CornerNet \cite{law2018cornernet}), one-stage models (YOLOv3 \cite{redmon2018yolov3}, RetinaNet \cite{lin2017focal}, FCOS \cite{tian2019fcos}), transformer-based model (DETR \cite{carion2020end}) and long-tailed learning model (LOCE \cite{feng2021exploring}). Shown in Fig. \ref{fig4}, our method outperforms others in several aspects. First, several advanced detectors such as Sparse R-CNN, YOLOv3, and LOCE missed abnormal cells, LSIL, in Fig. \ref{fig4}(d)(g)(k), which may result in low sensitivity in clinical cervical screening. Our model, equipped with inter-class instance relationship modeling (RIC and CIC), avoids these cases of missed detection. 
Then, our models achieved higher IoU detection scores compared with methods like Cascade R-CNN (Fig. \ref{fig4}(b)) and RepPoints (Fig. \ref{fig4}(e)). 
% In addition, Fig. \ref{fig5} highlights the superiority of the proposed model in detecting minority classes (e.g., SCC) and avoiding  domination of majority classes (e.g., HSIL).

% \begin{figure}[h!]
%     \centering
% \centerline{\includegraphics[width=0.5\textwidth]{Fig/F5.pdf}}
%     \caption{Detection results comparison between FRCNN (a) and our model (b) in minority class detection. (c) GT.}
%     \label{fig5}
% \end{figure}

\noindent\textbf{Quantitative Results.}
We provide the comprehensive quantitative result comparisons in Tab. \ref{tab4}, which shows that our holistic and historical comparison method achieves the best performance, with significant improvements of 13.8\% and 11.3\% in AP50 compared to the baseline with both R50 and R101 \cite{ren2015faster}. Our method effectively addresses class ambiguity, showing in improved detection accuracy for ambiguous classes, notably over 26\% growth in ASC-US. It also contributes to addressing class imbalance, with increased AP50 for minority classes, i.e., 9.8\% for SCC and 6.9\% for AGC-N. Our method outperforms two-stage and one-stage detectors, with a 2.9\% gain compared to the strongest two-stage method, Grid R-CNN \cite{lu2019grid}, and a 4.3\% improvement compared to the best one-stage detector, RepPoints \cite{yang2019reppoints}. Transformer-based models (e.g., DETR) achieve promising detection results, while they are limited by issues such as tiny targets and noise. Finally, a significant performance drop in majority classes leads to an overall performance decrease in long-tailed learning, i.e., LOCE \cite{feng2021exploring}, highlighting the effectiveness of our method in balancing the performance of different classes.

\subsection{Ablation Studies and Analysis}

\begin{table}[h!]
    \centering
    \small
    \caption{Ablation studies for parameters, temperature $\tau$ and memory bank size Q. \textcolor{red}{Red} denotes the best results.}
    \begin{tabular}{ccccc}
    \toprule[1pt]
        Parameter & AP50 $\uparrow$ & AP75 $\uparrow$& AP $\uparrow$& AR $\uparrow$ \\ 
        \cmidrule(lr){1-1}
        \cmidrule(lr){2-5}
        $\tau$ = 4 & 27.7 & 19.4 & 16.8 & 39.8  \\ 
        $\tau$ = 6 & \textcolor{red}{35.2} & \textcolor{red}{28.3} & \textcolor{red}{23.6} & \textcolor{red}{46.9} \\
        $\tau$ = 8 & 27.5 & 17.7 & 16.5 & 40.7 \\
        $\tau$ = 10 & 25.6 & 17 & 15.2 & 40.5 \\ 
        \cmidrule(lr){1-5}
        Q = 16 & 27.6 & 18.0 & 16.3 & 41.0 \\ 
        Q = 80 & \textcolor{red}{35.2} & \textcolor{red}{28.3} & \textcolor{red}{23.6} & \textcolor{red}{46.9} \\
        Q = 160 &  28.0 &  21.6 &  18.1 &  42.9 \\
        \bottomrule[1pt]
    \end{tabular}
    \vspace{-0.15in}
    \label{tab2}
\end{table}

\begin{table*}[]
    \centering
    % \scriptsize
    % \footnotesize
    \small
    % \tiny
    % \vspace{-0.1in}
    \caption{Further exploration for class imbalanced learning with focal \cite{lin2017focal} and seesaw loss \cite{wang2021seesaw}. \textcolor{red}{Red} denotes the best results.}
    % \vspace{-0.1in}
    % \setlength{\tabcolsep}{0.4mm}
    \begin{tabular}{cccccccccccccc}
    % \hline
    \toprule[1pt]
        Base & Loss & Ours & AP50 $\uparrow$ & AP75 $\uparrow$& AP $\uparrow$& AR $\uparrow$ & ASC-US & LSIL & ASC-H & HSIL & SCC & AGC & AGC-N \\ 
        \cmidrule(lr){1-3}
        \cmidrule(lr){4-7}
        \cmidrule(lr){8-14}
        \checkmark & ~ & ~ & 21.4 & 13.7 & 12.6 & 37.1 & 21.3 & 56.7 & 16.0 & 16.6 & 12.0 & 26.9 & 0.0 \\ 
        \cmidrule(lr){1-14}
        \checkmark & focal \cite{lin2017focal}& ~ & 18.5 & 13.9 & 11.9 & 46.8 & 17.3 & 53.4 & 15.3 & 14.0 & 1.3 & 25.7 & 2.6 \\
        \checkmark & focal \cite{lin2017focal}& \checkmark & \textcolor{red}{29.7} & \textcolor{red}{21.5} & \textcolor{red}{18.7} & \textcolor{red}{52.0} & \textcolor{red}{34.1} & \textcolor{red}{66.0} & \textcolor{red}{31.8} & \textcolor{red}{20.8} & \textcolor{red}{13.7} & \textcolor{red}{29.4} & \textcolor{red}{11.7} \\ 
        \cmidrule(lr){1-14}
        \checkmark & seesaw \cite{wang2021seesaw}& ~ & 33.5 & 22.9 & 20.9 & 54.2 & 29.9 & 63.9 & 28.9 & \textcolor{red}{22.7} & 36.5 & 34.8 & 17.6 \\
        \checkmark & seesaw \cite{wang2021seesaw}& \checkmark & \textcolor{red}{40.2} & \textcolor{red}{30.7} & \textcolor{red}{26.3} & \textcolor{red}{59.1} & \textcolor{red}{51.5} & \textcolor{red}{61.1} & \textcolor{red}{45.7} & 21.7 & \textcolor{red}{36.7} & \textcolor{red}{38.5} & \textcolor{red}{26.0} \\ 
        \bottomrule[1pt]
    \end{tabular}
        \label{tab3}
    \vspace{-0.05in}
\end{table*}

\begin{table*}[]
    \centering
    \small
    \caption{Effectiveness of proposed method in the CC-L dataset. \textcolor{red}{Red} denotes the performance changes compared to FRCNN.}
    \begin{tabular}{p{0.045\textwidth}p{0.04\textwidth}p{0.04\textwidth}p{0.04\textwidth}p{0.03\textwidth}p{0.04\textwidth}p{0.035\textwidth}p{0.04\textwidth}p{0.04\textwidth}p{0.04\textwidth}p{0.035\textwidth}p{0.04\textwidth}p{0.04\textwidth}p{0.04\textwidth}p{0.035\textwidth}p{0.04\textwidth}}
    \toprule[1pt]
        Method &ASCUS & LSIL& ASCH&HSIL&SCC&AGC&AGCN&ACTI&EMC&CC&Atrophy&FUNGI&Metapla&TV&AIS\\ 
        \cmidrule(lr){1-16}
         Baseline & 28.2 & 23.6& 25.5&3.3&1.5& 23.8&0.0& 73.7&39.5&44.3&18.1&10.5&4.6&22.2&0.0\\
       \cmidrule(lr){1-16}
       Ours & 31.1\textcolor{red}{\text{\tiny (+2.9)}} & 39.5\textcolor{red}{\text{\tiny (+15.9)}}& 31.0\textcolor{red}{\text{\tiny (+5.5)}}&6.8\textcolor{red}{\text{\tiny (+3.5)}}&13.5\textcolor{red}{\text{\tiny (+12.0)}} & 22.0\textcolor{red}{\text{\tiny (-1.8)}}&1.3\textcolor{red}{\text{\tiny (+1.3)}}& 86.3\textcolor{red}{\text{\tiny (+12.6)}}&47.6\textcolor{red}{\text{\tiny (+8.1)}}&39.3\textcolor{red}{\text{\tiny (-5.0)}}&33.6\textcolor{red}{\text{\tiny (+15.5)}}&19.6\textcolor{red}{\text{\tiny (+9.1)}}&18.9\textcolor{red}{\text{\tiny (+14.3)}}&24.4\textcolor{red}{\text{\tiny (+2.2)}}&14.6\textcolor{red}{\text{\tiny (+14.6)}}\\
    \bottomrule[1pt]
    \end{tabular}
    \label{tabe}
     \vspace{-0.15in}
\end{table*}

\noindent\textbf{Effectiveness of Instance Comparison.}
We perform ablation studies to investigate the effect of each proposed design. The comparison results can be seen in Tab. \ref{tab1}, with baseline (denoted as Base) and three components: RoI-level Instance Comparison (RIC), Box Augmentation (Aug), and Class-level Instance Comparison (CIC). Specifically, by adding RIC, we observe obvious performance improvements with 11.3\% and 8.9\% in AP50 and mean AP. For each class, we can see improvements both in majority (17.8\% for LSIL) and majority classes (7.3\% for SCC, 6.2\% for AGC, 3.0\% for AGC-N). Equipped with box augmentation for increasing deterministic instance number and diversity, the model gains further improvements of 1.8\% AP50 and 1.7\% AP. Class-level comparison aims to improve comparison capability via increasing frequencies of minority class, which yields improvements of 0.7\% AP50, notably 3.9\% for the minor class, AGC-N.

\noindent\textbf{Temperature Coefficient $\tau$.}
The temperature coefficient $\tau$ is a crucial parameter for penalizing hard negative samples more effectively, thereby directing the model's updates. To choose an appropriate value, we conduct a series of ablations with $\tau \in \{4,6,8,10\}$. As shown in Tab. \ref{tab2}, we achieve the highest detection metrics with 35.2\% AP50 and 23.6\% AP, which is set for the following experiments.

\noindent\textbf{Memory Bank Size Q.}
We employ memory banks with different sizes, and Tab. \ref{tab2} shows corresponding results for the task of cervical cell detection. The best performance is achieved when Q=80, with improvements observed for overall performance (exceeding 7.6\% AP50). This is because a too-small value (Q=16) cannot significantly benefit instance comparison learning, while a too-large value (Q=180) may lead to an empty queue for minority classes and full for majority classes in early iterations.

\noindent\textbf{Equipped with rebalancing loss.}
As a plug-and-play learning approach, we further explore our method by applying existing class re-balancing loss functions to the classification head. We replace the cross-entropy loss in vanilla FRCNN \cite{ren2015faster}, using Focal loss \cite{lin2017focal} and Seesaw loss \cite{wang2021seesaw}, shown in Tab. \ref{tab3}. They achieve noticeable improvements in minority classes, while sacrificing performance in majority classes. When equipped with the proposed instance comparison approach, both Focal loss and Seesaw loss achieve consistent improvements, namely 11.2\% AP50 for Focal loss and 6.7\% AP50 for Seesaw loss.

\noindent\textbf{Experiments on the large-scale cytology dataset.} Moreover, we conduct experiments on CC-L, demonstrating the effectiveness of our approach on datasets with increasingly larger and more complex structures. As shown in Tab.\ref{tabe}, we achieve 28.6\% AP50 and 16.6\% AP results with significant performance improvements in almost all categories. Specifically, we can see the effectiveness of our method for addressing class ambiguity, as in typical ambiguity categories, ASC-US (2.9\%) and ASC-H (5.5\%). Moreover, performance compensation is significantly highlighted in minor categories, as from 0.0\% to 14.6\% in AIS, and from 1.5\% to 13.5\% in SCC.

\section{Conclusion}
In this work, we investigated the intrinsical issues of class ambiguity and imbalance in cervical cell detection. To jointly address these issues, we propose a novel instance comparison approach with holistic and historical cell comparison at both the RoI-level and class-level, together with a confident sample selection-based memory bank for compensating the contribution from minority class instances.
Our experiments on two large-scale cervical cytology datasets demonstrate the effectiveness of our approach. The performance improvements of minority classes, e.g., AGC-N, remain to be further explored.

\section{Acknowledgments}
This work was supported by the National Natural Science Foundation of China (No. 62202403), Hong Kong Innovation and Technology Fund (Project No. PRP/034/22FX), Shenzhen Science and Technology Innovation Committee Fund (Project No. KCXFZ20230731094059008) and the Project of Hetao Shenzhen-Hong Kong Science and Technology Innovation Cooperation Zone (HZQB-KCZYB-2020083).

{\small
\bibliographystyle{IEEEtranS}
\bibliography{egbib}

% Generated by IEEEtranS.bst, version: 1.14 (2015/08/26)
\begin{thebibliography}{10}
\providecommand{\url}[1]{#1}
\csname url@samestyle\endcsname
\providecommand{\newblock}{\relax}
\providecommand{\bibinfo}[2]{#2}
\providecommand{\BIBentrySTDinterwordspacing}{\spaceskip=0pt\relax}
\providecommand{\BIBentryALTinterwordstretchfactor}{4}
\providecommand{\BIBentryALTinterwordspacing}{\spaceskip=\fontdimen2\font plus
\BIBentryALTinterwordstretchfactor\fontdimen3\font minus
  \fontdimen4\font\relax}
\providecommand{\BIBforeignlanguage}[2]{{%
\expandafter\ifx\csname l@#1\endcsname\relax
\typeout{** WARNING: IEEEtranS.bst: No hyphenation pattern has been}%
\typeout{** loaded for the language `#1'. Using the pattern for}%
\typeout{** the default language instead.}%
\else
\language=\csname l@#1\endcsname
\fi
#2}}
\providecommand{\BIBdecl}{\relax}
\BIBdecl

\bibitem{cai2018cascade}
Z.~Cai and N.~Vasconcelos, ``Cascade r-cnn: Delving into high quality object
  detection,'' in \emph{Proceedings of the IEEE conference on computer vision
  and pattern recognition}, 2018, pp. 6154--6162.

\bibitem{carion2020end}
N.~Carion, F.~Massa, G.~Synnaeve, N.~Usunier, A.~Kirillov, and S.~Zagoruyko,
  ``End-to-end object detection with transformers,'' in \emph{European
  conference on computer vision}.\hskip 1em plus 0.5em minus 0.4em\relax
  Springer, 2020, pp. 213--229.

\bibitem{chai2022deep}
Z.~Chai, L.~Luo, H.~Lin, H.~Chen, A.~Han, and P.-A. Heng, ``Deep
  semi-supervised metric learning with dual alignment for cervical cancer cell
  detection,'' in \emph{2022 IEEE 19th International Symposium on Biomedical
  Imaging (ISBI)}.\hskip 1em plus 0.5em minus 0.4em\relax IEEE, 2022, pp. 1--5.

\bibitem{feng2021exploring}
C.~Feng, Y.~Zhong, and W.~Huang, ``Exploring classification equilibrium in
  long-tailed object detection,'' in \emph{Proceedings of the IEEE/CVF
  International conference on computer vision}, 2021, pp. 3417--3426.

\bibitem{jiang2020geometry}
H.~Jiang, S.~Li, W.~Liu, H.~Zheng, J.~Liu, and Y.~Zhang, ``Geometry-aware cell
  detection with deep learning,'' \emph{Msystems}, vol.~5, no.~1, pp. 10--1128,
  2020.

\bibitem{jiang2023donet}
H.~Jiang, R.~Zhang, Y.~Zhou, Y.~Wang, and H.~Chen, ``Donet: Deep de-overlapping
  network for cytology instance segmentation,'' in \emph{Proceedings of the
  IEEE/CVF Conference on Computer Vision and Pattern Recognition}, 2023, pp.
  15\,641--15\,650.

\bibitem{jiang2022deep}
H.~Jiang, Y.~Zhou, Y.~Lin, R.~C. Chan, J.~Liu, and H.~Chen, ``Deep learning for
  computational cytology: A survey,'' \emph{Medical Image Analysis}, p. 102691,
  2022.

\bibitem{law2018cornernet}
H.~Law and J.~Deng, ``Cornernet: Detecting objects as paired keypoints,'' in
  \emph{Proceedings of the European conference on computer vision (ECCV)},
  2018, pp. 734--750.

\bibitem{liang2022exploring}
Y.~Liang, S.~Feng, Q.~Liu, H.~Kuang, J.~Liu, L.~Liao, Y.~Du, and J.~Wang,
  ``Exploring contextual relationships for cervical abnormal cell detection,''
  \emph{arXiv preprint arXiv:2207.04693}, 2022.

\bibitem{lin2021dual}
H.~Lin, H.~Chen, X.~Wang, Q.~Wang, L.~Wang, and P.-A. Heng, ``Dual-path network
  with synergistic grouping loss and evidence driven risk stratification for
  whole slide cervical image analysis,'' \emph{Medical Image Analysis},
  vol.~69, p. 101955, 2021.

\bibitem{lin2017focal}
T.-Y. Lin, P.~Goyal, R.~Girshick, K.~He, and P.~Doll{\'a}r, ``Focal loss for
  dense object detection,'' in \emph{Proceedings of the IEEE international
  conference on computer vision}, 2017, pp. 2980--2988.

\bibitem{liu2022sample}
M.~Liu, X.~Li, X.~Gao, J.~Chen, L.~Shen, and H.~Wu, ``Sample hardness based
  gradient loss for long-tailed cervical cell detection,'' in \emph{Medical
  Image Computing and Computer Assisted Intervention--MICCAI 2022: 25th
  International Conference, Singapore, September 18--22, 2022, Proceedings,
  Part II}.\hskip 1em plus 0.5em minus 0.4em\relax Springer, 2022, pp.
  109--119.

\bibitem{lu2019grid}
X.~Lu, B.~Li, Y.~Yue, Q.~Li, and J.~Yan, ``Grid r-cnn,'' in \emph{Proceedings
  of the IEEE/CVF Conference on Computer Vision and Pattern Recognition}, 2019,
  pp. 7363--7372.

\bibitem{nayar2015bethesda}
R.~Nayar and D.~C. Wilbur, \emph{The Bethesda system for reporting cervical
  cytology: definitions, criteria, and explanatory notes}.\hskip 1em plus 0.5em
  minus 0.4em\relax Springer, 2015.

\bibitem{redmon2018yolov3}
J.~Redmon and A.~Farhadi, ``Yolov3: An incremental improvement,'' \emph{arXiv
  preprint arXiv:1804.02767}, 2018.

\bibitem{ren2015faster}
S.~Ren, K.~He, R.~Girshick, and J.~Sun, ``Faster r-cnn: Towards real-time
  object detection with region proposal networks,'' \emph{Advances in neural
  information processing systems}, vol.~28, 2015.

\bibitem{sun2021sparse}
P.~Sun, R.~Zhang, Y.~Jiang, T.~Kong, C.~Xu, W.~Zhan, M.~Tomizuka, L.~Li,
  Z.~Yuan, C.~Wang \emph{et~al.}, ``Sparse r-cnn: End-to-end object detection
  with learnable proposals,'' in \emph{Proceedings of the IEEE/CVF conference
  on computer vision and pattern recognition}, 2021, pp. 14\,454--14\,463.

\bibitem{sung2021global}
H.~Sung, J.~Ferlay, R.~L. Siegel, M.~Laversanne, I.~Soerjomataram, A.~Jemal,
  and F.~Bray, ``Global cancer statistics 2020: Globocan estimates of incidence
  and mortality worldwide for 36 cancers in 185 countries,'' \emph{CA: a cancer
  journal for clinicians}, vol.~71, no.~3, pp. 209--249, 2021.

\bibitem{tian2019fcos}
Z.~Tian, C.~Shen, H.~Chen, and T.~He, ``Fcos: Fully convolutional one-stage
  object detection,'' in \emph{Proceedings of the IEEE/CVF international
  conference on computer vision}, 2019, pp. 9627--9636.

\bibitem{wang2021seesaw}
J.~Wang, W.~Zhang, Y.~Zang, Y.~Cao, J.~Pang, T.~Gong, K.~Chen, Z.~Liu, C.~C.
  Loy, and D.~Lin, ``Seesaw loss for long-tailed instance segmentation,'' in
  \emph{Proceedings of the IEEE/CVF conference on computer vision and pattern
  recognition}, 2021, pp. 9695--9704.

\bibitem{yang2019reppoints}
Z.~Yang, S.~Liu, H.~Hu, L.~Wang, and S.~Lin, ``Reppoints: Point set
  representation for object detection,'' in \emph{Proceedings of the IEEE/CVF
  international conference on computer vision}, 2019, pp. 9657--9666.

\end{thebibliography}
}
\end{document}